\documentclass[letterpaper]{article} % DO NOT CHANGE THIS
\usepackage{aaai2027} % DO NOT CHANGE THIS
\nocopyright
\usepackage[hyphens]{url} % DO NOT CHANGE THIS
\usepackage{graphicx} % DO NOT CHANGE THIS
\usepackage{natbib} % DO NOT CHANGE THIS AND DO NOT ADD OPTIONS
\usepackage{caption} % DO NOT CHANGE THIS AND DO NOT ADD OPTIONS
\usepackage{amsmath}
\usepackage{amsfonts}
\usepackage{amssymb}
\usepackage{booktabs}
\usepackage{booktabs}
\usepackage{multirow}
\usepackage{array}
\usepackage{natbib}
\usepackage{caption}
\title{Beyond Response Scoring: An AI-aware Cognitive Supervision Framework for Automated Mathematical Short Answer Grading}

\author{
    Shengran Guo,
    Min Chi,
    Feifei Qiao,
    Varad Patwardhan
}

\affiliations{
    Department of Computer Science\\
    North Carolina State University\\
    Raleigh, NC, USA\\
}

\begin{document}

\maketitle

\begin{abstract}
    \vspace{-2pt}
Transformer-based language models have achieved remarkable success in automated short answer grading (ASAG).
However, current methods in ASAG struggle to comprehend, reason about and evaluate complex mathematical answers involving symbols and formulas.
These score-only methods fail to generalize across diverse responses and skill states associated with identical scores.
Moreover, they ignore the potential upward bias introduced by generative Artificial Intelligence (AI) usage in assessing students' performance and knowledge mastery. To address these limitations, we propose \textbf{AICG}
(\textbf{AI}-aware \textbf{C}ognitive-supervised \textbf{G}rading), a
two-branch multi-task framework. The first branch combines RoBERTa-based regression, ordinal, and classification heads for individual-answer grading. The second branch uses an expert-designed Q-matrix to associate responses with assessed skills and jointly predicts G-DINA-derived skill mastery and learning gain. Through the shared encoder, these skill-level targets provide cognitive supervision for answer grading beyond score labels. AICG further uses a Gen-AI detector, trained through supervised contrastive pretraining followed by linear classification, to estimate assistance probabilities and convert them into reliability weights, attenuating unreliable cognitive evidence during learning assessment. We conduct experiments on 49,402 instructor-graded responses spanning 19
academic semesters. Relative to
RoBERTa regression, AICG reduces mean absolute error (MAE) by 50.3\% and
improves quadratic weighted kappa (QWK) from 0.798 to 0.927. Ablations
identify Q-matrix-guided cognitive supervision as the primary source of
improvement, while AI-aware weighting further reduces overall MAE by 4.2\% over the
unweighted cognitive model and consistently improves MAE and QWK across the four most recent post-Gen-AI semesters.

\end{abstract}

\section{Introduction}
\label{sec:introduction}
% \textbf{Automated grading of mathematical short answers} sits at a demanding intersection of \emph{natural language understanding}, \emph{symbolic reasoning}, and \emph{cognitive assessment}: responses interleave prose and derivations, partial credit depends on which reasoning principles were demonstrated rather than merely whether the final answer is correct. Moreover, in the era of generative AI, the same fluent, high-scoring response may reflect a student's genuine mastery or conceal a borrowed solution, making the accuracy and integrity of scalable grading more consequential than ever. 
% Existing work approaches this problem from two complementary directions: \emph{individual-answer grading}, which maps each response directly to a score, and \emph{Q-matrix-based skill modeling}, which grounds assessment in an explicit cognitive structure. Each line of work has made substantial progress, yet each faces limitations that the other does not address, and neither contends with the reliability threat introduced by generative AI.

\textbf{Automated grading of mathematical short answers} sits at the
intersection of \emph{natural language understanding}, \emph{symbolic
reasoning}, and \emph{cognitive assessment}: responses interleave prose
and derivations, while partial credit depends on demonstrated reasoning
rather than only the final answer. With generative AI, the same fluent,
high-scoring response may reflect genuine mastery or a borrowed
solution, increasing the importance of accurate and reliable grading.
Existing work approaches this problem from two complementary
directions: \emph{individual-answer grading}, which maps each response
to a score, and \emph{Q-matrix-based skill modeling}, which grounds
assessment in an explicit cognitive structure. Both have made
substantial progress, but neither addresses their complementary
limitations or the reliability threat introduced by generative AI.

\textbf{Individual-answer grading.} In automated short-answer grading
(ASAG), pretrained language models and LLMs encode response semantics
with substantially less manual feature engineering
\cite{burrows2015automated,latif2024finetuning,mello2025automatic,
chu2025gradeopt}. However, mathematical responses interleave
natural-language explanations, symbolic formulas, and incomplete
derivations, while partial credit may depend on multiple reasoning
principles. Current language models remain unreliable in
domain-specific symbolic and compositional reasoning
\cite{latif2024finetuning,jiang2024short,mello2025automatic} and may
misinterpret valid intermediate or mathematically equivalent work.
More fundamentally, supervision is compressed into a scalar label:
identical scores can represent different reasoning and skill states,
limiting generalization to unseen solution patterns.
Figure~\ref{fig:partial_credit_example} illustrates this difficulty:
the student correctly applies conditional independence twice but stops
before Bayes' rule; the score of $0.60$ records the earned credit but
not the underlying reasoning pattern.

\textbf{Q-matrix-based skill modeling.} A complementary perspective
explicitly represents the cognitive structure of assessment. A Q-matrix
associates items with skills, while cognitive-diagnosis models such as
G-DINA infer mastery from response patterns
\cite{delaTorre2011GDINA}. Student and question representations
complement answer features \cite{zhang2020going}, and neural
cognitive-diagnosis models demonstrate the value of explicit skill
structure
\cite{wang2020neuralcd,liu2024inductive,qian2024orcdf,
gu2025revisiting}. Figure~\ref{fig:partial_credit_example} illustrates
this value: the missing Bayes step indicates weak Bayesian reasoning,
while the Q-matrix links this evidence across items assessing the same
skill. However, these models rely on binary item-response records and
leave the semantic evidence in free-text solutions unused. Generative
AI further complicates assessment because an assisted answer may earn
its assigned grade while overstating independent mastery. Although
recent detectors identify machine-generated text
\cite{hans2024binoculars,bao2024fastdetectgpt}, hard filtering remains
vulnerable to rewriting and distribution shift and may disproportionately
affect non-native writers
\cite{sadasivan2025can,liang2023detectors}, making binary removal
unsuitable.

\begin{figure}[t]
\centering
\setlength{\fboxsep}{2pt}

\definecolor{language}{RGB}{31,78,121}
\definecolor{earned}{RGB}{0,105,62}
\definecolor{missing}{RGB}{176,45,35}
\definecolor{panelgray}{RGB}{247,247,247}
\definecolor{referencebg}{RGB}{244,248,252}
\definecolor{earnedbg}{RGB}{242,250,246}
\definecolor{missingbg}{RGB}{253,244,243}

\fcolorbox{black}{panelgray}{%
\begin{minipage}{0.96\columnwidth}
\fontsize{6.5}{7.3}\selectfont
\raggedright
\setlength{\parindent}{0pt}
\setlength{\parskip}{0pt}
\setlength{\mathsurround}{0pt}

\newcommand{\answerblock}[4]{%
\setlength{\fboxsep}{1.2pt}%
\noindent\fcolorbox{#1}{#2}{%
\begin{minipage}{\dimexpr\linewidth-2\fboxsep-2\fboxrule\relax}
\raggedright
\hfill\textcolor{#1}{\textbf{#3}}\par
#4
\end{minipage}}\par}

%----------------------------------------------------------
% Question
%----------------------------------------------------------
\textbf{Question.}
Given $p(O)=0.5$, suppose $A$ and $B$ are conditionally independent given
$O$ and $\bar O$. Let $p(A\mid O)=p(\bar A\mid\bar O)=0.8$,
$p(B\mid O)=0.75$, and $p(B\mid\bar O)=0.5$.
Determine $p(O\mid A\cap B)$.

\vspace{1pt}\hrule\vspace{1pt}

%----------------------------------------------------------
% Correct answer
%----------------------------------------------------------
\textbf{Reference Answer}\hfill\textbf{Principle/Weight}\par

\answerblock{language}{referencebg}{Definition of Conditional Independent Events (0.3)}{%
$p(A\cap B\mid O)=p(A\mid O)*P(B\mid O)=0.8*0.75=0.6.$%
}

\vspace{0.5pt}

\answerblock{language}{referencebg}{Definition of Conditional Independent Events (0.3)}{%
$p(A\cap B\mid{\sim}O)=p(A\mid{\sim}O)*P(B\mid{\sim}O)=(1-0.8)*0.5=0.10.$%
}

\vspace{0.5pt}

\answerblock{language}{referencebg}{Bayes’s Rule (0.4)}{%
$\displaystyle
p(O\mid A\cap B)=
\frac{p(A\cap B\mid O)*p(O)}
{p(A\cap B\mid O)*p(O)+p(A\cap B\mid{\sim}O)*p({\sim}O)}
$\par
$=0.6*0.5/(0.6*0.5+0.1*0.5)=0.857143.$%
}

\vspace{1pt}\hrule\vspace{1pt}

%----------------------------------------------------------
% Student answer
%----------------------------------------------------------
\textbf{Student Answer}\hfill\textbf{Total score: 0.60}\par

\answerblock{earned}{earnedbg}{Definition of Conditional Independent Events: 1}{%
{
\emph{conditionally independent events:}
$p(A\cap B\mid O)=p(A\mid O)*p(B\mid O)$\par
$\texttt{--->}\;p(A\cap B\mid O)=0.8*0.75=0.6$
}%
}
\vspace{0.5pt}

\answerblock{earned}{earnedbg}{Definition of Conditional Independent Events: 1}{%
{
$p(A\cap B\mid{\sim}O)=p(A\mid{\sim}O)*p(B\mid{\sim}O)$
$\texttt{--->}\;0.2*0.5=0.1$
}%
}
\vspace{0.5pt}

\answerblock{missing}{missingbg}{Bayes’s Rule: 0}{%
{\emph{I don't know how to do the rest}}%
}

\vspace{1pt}\hrule\vspace{1pt}

%----------------------------------------------------------
% Score summary
%----------------------------------------------------------

\textbf{Weighted score:} $0.3(1)+0.3(1)+0.4(0)=0.60$.\par

\end{minipage}}

\vspace{-2pt}
\caption{Alignment between weighted reference principles and partial-credit
evidence. Blue, green, and red boxes denote reference steps,
demonstrated principles, and the missing application of Bayes's rule,
respectively.}
\label{fig:partial_credit_example}
\vspace{-1.5em}
\end{figure}
We propose \textbf{AICG}
(\textbf{AI}-aware \textbf{C}ognitive-supervised \textbf{G}rading), a
two-branch multi-task framework. The first branch applies regression, ordinal,
and classification heads to a shared RoBERTa representation
\cite{liu2019roberta}, capturing complementary continuous, ordered, and
categorical grading signals. The second branch uses the Q-matrix to aggregate
skill-relevant responses and predicts G-DINA-derived pretest mastery, posttest
mastery, and learning gain. Joint optimization transfers this cognitive
supervision to the shared response encoder without feeding predicted grades
into the cognitive branch. AICG further trains a Gen-AI detector through
supervised contrastive pretraining \cite{khosla2020supervised} followed by
linear classification. The estimated assistance probabilities are converted
into soft reliability weights applied only to cognitive evidence, preserving
the grading treatment of each response. Our contributions are threefold:
\begin{itemize}
    \item AICG is the first Transformer-based ASAG framework to jointly
    optimize answer grading with Q-matrix-aligned mastery and learning-gain
    supervision.

    \item We introduce AI-aware reliability weighting that attenuates
    potentially assisted cognitive evidence without discarding responses or
    altering their grades.

    \item AICG outperforms five strong baselines in a ten-year
    expanding-window evaluation while maintaining robust cross-semester generalization.
\end{itemize}

\section{Methodology}
\label{sec:method}
\begin{figure*}[t]
    \centering
    \includegraphics[width=0.9\textwidth]{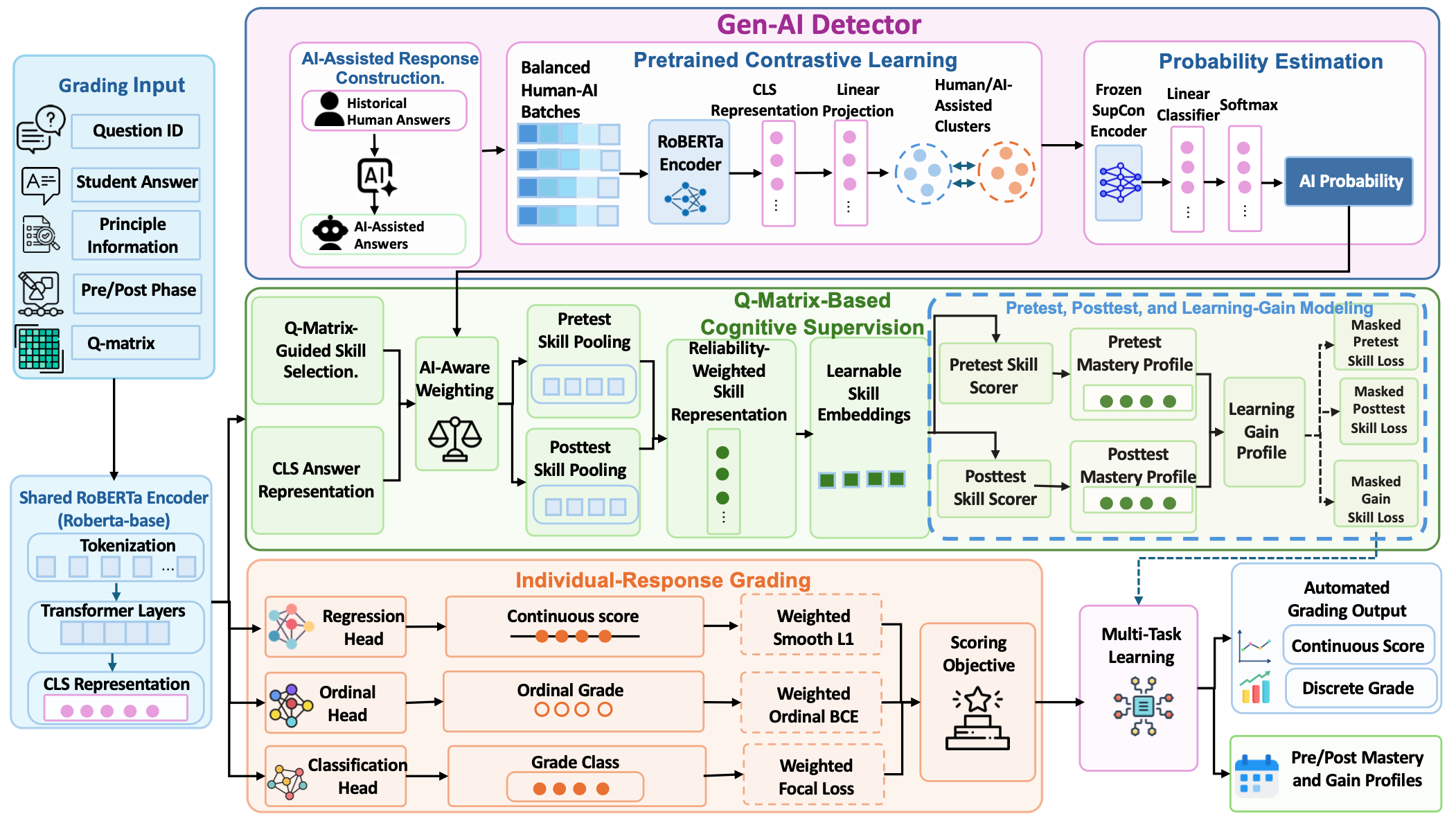}
   \caption{Overview of AICG. The Gen-AI Detector supplies AI-assistance
probabilities only to the Q-Matrix-Based Cognitive Supervision branch.
The shared RoBERTa encoder supports parallel cognitive-supervision and
individual-response-grading branches. Solid and dashed arrows denote
prediction and supervision flows, respectively.}
    \label{fig:aicg_framework}
    \vspace{-1em}
\end{figure*}

%%%%%%%%%%%%%%%%%%%%%%%%%%%%%%%%%%%%%%%%%%%%%%%%%%%%%%%%%%%%
\vspace{-2pt}
Figure~\ref{fig:aicg_framework} presents AICG, which comprises two
major components. The first is the \emph{Gen-AI Detector}, which
assesses AI involvement in responses ranging from partially assisted
to fully generated and assigns each response a continuous AI-assistance
probability. The second is a multi-task grading component built on a
separate RoBERTa encoder shared by two branches. The
\emph{Individual-Response Grading} branch predicts continuous, ordinal,
and categorical grades from each response's semantic representation.
The \emph{Q-Matrix-Based Cognitive Supervision} branch combines Q-matrix-guided skill selection with
AI-aware evidence weighting to predict pretest mastery, posttest mastery, and learning gain, thereby supervising grading with cognitive information. The
Gen-AI Detector is used only to adjust the reliability of cognitive
evidence. The two branches retain independent heads while jointly updating the shared RoBERTa encoder.
%Predicted grades never enter the cognitive branch.

\vspace{-2pt}
\subsection{Problem Formulation}
\label{sec:problem}
%%%%%%%%%%%%%%%%%%%%%%%%%%%%%%%%%%%%%%%%%%%%%%%%%%%%%%%%%%%
\vspace{-2pt}
Let $\mathcal{S}$ denote the student set, with $s$ indexing a student;
let $\mathcal{T}=\{\mathrm{pre},\mathrm{post}\}$ denote the assessment
phases, with $t\in\mathcal{T}$ indexing a phase; and let
$\mathcal{K}=\{1,\ldots,K\}$ denote the cognitive-skill set, with $k$
indexing a skill and $K=10$. For student $s$ in phase $t$,
$\mathcal{R}_{s}^{t}$ denotes the observed-response set, with $r$
indexing a response.

For response $r$ from student $s$, $x_{s}^{r}$ denotes the textual
input and $y_{s}^{r}\in[0,1]$ the normalized instructor-assigned
score. The input $x_{s}^{r}$ concatenates the question identifier,
student answer, and principle metadata using separator tokens.
The Q-matrix vector $\mathbf{q}_{s}^{r}\in\{0,1\}^{K}$ associates the
response with assessed skills: $q(s,r,k)\in\{0,1\}$ indicates whether
the corresponding item assesses skill $k$. The assessment phase and
Q-matrix vector remain structured inputs rather than being appended to
the response text.

At the response level, $\hat{y}_{s}^{r}\in[0,1]$ denotes the predicted
continuous score. The ground-truth and predicted ordered categories
are denoted by $d_{s}^{r}\in\{0,1,2,3,4\}$ and
$\hat{d}_{s}^{r}\in\{0,1,2,3,4\}$, respectively. The ground-truth
category is derived from $y_{s}^{r}$ using boundaries $0.125$,
$0.375$, $0.625$, and $0.875$.

At the student--skill level,
$\widetilde{c}(s,k,t)\in[0,1]$ denotes the reliability-adjusted G-DINA
mastery target, and $\hat{c}(s,k,t)\in[0,1]$ its prediction. The
reliability-adjusted learning-gain target
$\widetilde{g}_{s}^{k}\in[-1,1]$ is the difference between the
corresponding posttest and pretest mastery targets, while
$\hat{g}_{s}^{k}\in[-1,1]$ denotes its prediction. AICG is supervised
by $\widetilde{c}(s,k,t)$ and $\widetilde{g}_{s}^{k}$, which are
derived from AI-aware, Q-matrix-aligned response evidence.

\subsection{Gen-AI Detector}
\label{sec:ai_detection}
%%%%%%%%%%%%%%%%%%%%%%%%%%%%%%%%%%%%%%%%%%%%%%%%%%%%%%%%%%%%
\vspace{-2pt}
An AI-assisted response may provide unreliable evidence of independent
mastery. We therefore train the Gen-AI Detector separately to estimate
continuous AI-assistance probabilities.

\paragraph{AI-Assisted Response Construction.}
Complete LLM rewriting can introduce stylistic artifacts that encourage
shortcut learning. We instead mask 30\%--50\% of each authentic response
and use an offline DeepSeek model to reconstruct the masked tokens,
preserving 50\%--70\% of the student text. Original responses are labeled
human-written; partially completed and fully generated correct responses
form the AI-assisted class. Each source response and its variants remain
in the same partition to prevent leakage. Classes are matched by item,
score category, and length, and generator transfer is evaluated through
leave-one-generator-out splits. Further details are provided in Appendix.

\paragraph{Supervised Contrastive Pretraining.}

For response $r$, $\mathbf{v}_r$ represents the textual input to the
Gen-AI Detector, and $a_r\in\{0,1\}$ represents its source label, with $a_r=0$
for human-written text and $a_r=1$ for partially assisted or fully
generated text. Training uses balanced mini-batches with equal numbers
of responses from both classes.

The Gen-AI detector employs a RoBERTa encoder
$\operatorname{RoBERTa}_{\mathrm{det}}$, where $\mathrm{[CLS]}$
denotes the final hidden state of the sequence-initial classification
token. The learnable projection matrix and bias are represented by
$\mathbf{W}_{\mathrm{proj}}$ and $\mathbf{b}_{\mathrm{proj}}$,
respectively. The projected representation $\mathbf{u}_r$ is calculated
as:
\vspace{-2pt}
\begin{equation}
\mathbf{u}_r
=
\mathbf{W}_{\mathrm{proj}}
\operatorname{RoBERTa}_{\mathrm{det}}
\left(\mathbf{v}_r\right)_{\mathrm{[CLS]}}
+
\mathbf{b}_{\mathrm{proj}}.
\label{eq:detector_projection}
\end{equation}

The normalized contrastive embedding is denoted by $\mathbf{e}_r$, and
$\|\cdot\|_2$ denotes the Euclidean norm. It is computed as:
\vspace{-2pt}
\begin{equation}
\mathbf{e}_r
=
\frac{\mathbf{u}_r}{\|\mathbf{u}_r\|_2}.
\label{eq:detector_normalization}
\end{equation}

Let $\mathcal{B}$ denote the response indices in a balanced mini-batch.
For anchor $r\in\mathcal{B}$, $\mathcal{P}(r)$ contains the non-anchor
responses sharing label $a_r$; $r^{+}$ indexes a response in
$\mathcal{P}(r)$, and $r'$ indexes a non-anchor response in
$\mathcal{B}$. With temperature $\tau=0.05$, the supervised contrastive
objective $\mathcal{L}_{\mathrm{con}}$ encourages within-class
compactness and between-class separation. It is formulated as:
\vspace{-2pt}
\begin{equation}
\mathcal{L}_{\mathrm{con}}
=
-\frac{1}{|\mathcal{B}|}
\sum_{r\in\mathcal{B}}
\frac{1}{|\mathcal{P}(r)|}
\sum_{r^{+}\in\mathcal{P}(r)}
\log
\frac{
\exp\!\left(
\mathbf{e}_{r}^{\top}\mathbf{e}_{r^{+}}/\tau
\right)
}{
\displaystyle
\sum_{r'\in\mathcal{B}\setminus\{r\}}
\exp\!\left(
\mathbf{e}_{r}^{\top}\mathbf{e}_{r'}/\tau
\right)
}.
\vspace{-2pt}
\label{eq:supcon}
\end{equation}

\vspace{-2pt}
\paragraph{Probability Estimation.}

After contrastive pretraining, the Gen-AI Detector encoder and
projection layer are frozen. A two-class linear classifier,
parameterized by $\mathbf{W}_{\mathrm{cls}}$ and
$\mathbf{b}_{\mathrm{cls}}$, is trained on the frozen representations
$\mathbf{u}_r$ using cross-entropy loss.

Probability calibration uses the group-disjoint validation set
$\mathcal{V}_{\mathrm{cal}}$, whose size is
$|\mathcal{V}_{\mathrm{cal}}|$. A candidate temperature is represented
by $T>0$, and $T^{*}$ is the value selected by minimizing the average
negative log-likelihood. The subscript $a_r$ selects the probability
assigned to the ground-truth source label of response $r$. The optimal
temperature is calculated as:
\vspace{-2pt}
\begin{equation}
\begin{aligned}
T^{*}
&=
\underset{T>0}{\arg\min}\;
-\frac{1}{|\mathcal{V}_{\mathrm{cal}}|}
\sum_{r\in\mathcal{V}_{\mathrm{cal}}}
\\[-2pt]
&\quad
\log\!\Bigg[
\operatorname{Softmax}
\left(
\frac{
\mathbf{W}_{\mathrm{cls}}\mathbf{u}_r+
\mathbf{b}_{\mathrm{cls}}
}{T}
\right)_{a_r}
\Bigg].
\end{aligned}
\label{eq:detector_temperature}
\end{equation}

The calibrated AI-assistance probability for response $r$ is denoted
by $p_r^{\mathrm{AI}}\in[0,1]$. Because $a_r=1$ represents partially
assisted or fully generated text, this probability is computed as the
class-1 output:
\vspace{-2pt}
\begin{equation}
p_r^{\mathrm{AI}}
=
\operatorname{Softmax}
\left(
\frac{
\mathbf{W}_{\mathrm{cls}}\mathbf{u}_r+
\mathbf{b}_{\mathrm{cls}}
}{T^{*}}
\right)_{1}.
\label{eq:detector_probability}
\end{equation}

For the grading dataset, $p^{\mathrm{AI}}(s,r)$ denotes the calibrated
AI-assistance probability of response $r$ from student $s$.
Calibration is evaluated using expected calibration error and the
Brier score. Grading-test responses are excluded from Gen-AI Detector
training and temperature estimation.

\subsection{Individual-Response Grading}
\label{sec:aicg}
%%%%%%%%%%%%%%%%%%%%%%%%%%%%%%%%%%%%%%%%%%%%%%%%%%%%%%%%%%%%
\vspace{-2pt}
As shown in Figure~\ref{fig:aicg_framework}, AICG employs a RoBERTa
encoder shared with the cognitive branch but separate from the Gen-AI
Detector. For input $x_s^r$, the encoder produces
$\mathbf{h}_s^r\in\mathbb{R}^{D}$, the dropout-regularized final hidden
state of the sequence-initial $\mathrm{[CLS]}$ token, where $D$ is the
hidden dimension. Regression, ordinal, and categorical heads operate
independently on $\mathbf{h}_s^r$. The regression head uses a sigmoid
function $\sigma(\cdot)$ to produce the continuous score
$\hat{y}_s^r$. The ordinal head produces
$\mathbf{o}_s^r\in\mathbb{R}^{4}$, where $o(s,r,j)$ denotes the logit
for threshold $j\in\{0,\ldots,3\}$. The categorical head produces
$\boldsymbol{\xi}_s^r\in\mathbb{R}^{5}$ for the five grade categories.

For category $d\in\{0,\ldots,4\}$, $\alpha_d$ denotes its training
weight. We set
$(\alpha_0,\ldots,\alpha_4)=(1.0,2.0,2.8,2.0,0.7)$ and assign
$\eta_s^r=\alpha_{d_s^r}$ to response $r$ from student $s$. The
mini-batch $\mathcal{B}_{\mathrm{grade}}$ contains student--response
pairs, and its total weight is
$\Omega_{\mathrm{grade}}
=\sum_{(s,r)\in\mathcal{B}_{\mathrm{grade}}}\eta_s^r$.

The regression loss $\mathcal{L}_{\mathrm{reg}}$ uses
$\operatorname{SmoothL1}_{\beta}$, which is quadratic below
$\beta=0.10$ and linear otherwise. Its category-weighted average is
calculated as:

\begin{equation}
\mathcal{L}_{\mathrm{reg}}
=
\frac{
\displaystyle
\sum_{(s,r)\in\mathcal{B}_{\mathrm{grade}}}
\eta_s^r
\operatorname{SmoothL1}_{\beta}
\left(\hat{y}_s^r,y_s^r\right)
}{
\Omega_{\mathrm{grade}}
}.
\label{eq:regression_loss}
\end{equation}

For threshold $j$, the ordinal target is
$\kappa(s,r,j)=\mathbb{1}[d_s^r>j]$, where
$\mathbb{1}[\cdot]$ denotes the indicator function. The positive and
negative training counts are $N_j^{+}$ and $N_j^{-}$, respectively,
and the positive-class weight is
$\nu_j=\operatorname{clip}
(N_j^{-}/\max(N_j^{+},1),0.15,5.0)$.
Using binary cross-entropy with logits
$\operatorname{BCE}_{\nu_j}$, the ordinal loss
$\mathcal{L}_{\mathrm{ord}}$ is calculated as:
\vspace{-2pt}
\begin{equation}
\mathcal{L}_{\mathrm{ord}}
=
\frac{
\displaystyle
\sum_{(s,r)\in\mathcal{B}_{\mathrm{grade}}}
\frac{\eta_s^r}{4}
\sum_{j=0}^{3}
\operatorname{BCE}_{\nu_j}
\left(o(s,r,j),\kappa(s,r,j)\right)
}{
\Omega_{\mathrm{grade}}
}.
\label{eq:ordinal_loss}
\end{equation}

The categorical probabilities are
$\boldsymbol{\pi}_s^r=
\operatorname{Softmax}(\boldsymbol{\xi}_s^r)$, where
$\pi(s,r,d)$ denotes the probability of category $d$. With focal
parameter $\gamma=2$, the category-weighted classification loss
$\mathcal{L}_{\mathrm{cls}}$ is defined as:
\vspace{-2pt}
\begin{equation}
\mathcal{L}_{\mathrm{cls}}
=
-\frac{
\displaystyle
\sum_{(s,r)\in\mathcal{B}_{\mathrm{grade}}}
\eta_s^r
\left(1-\pi(s,r,d_s^r)\right)^{\gamma}
\log \pi(s,r,d_s^r)
}{
\Omega_{\mathrm{grade}}
}.
\label{eq:classification_loss}
\end{equation}

The three losses form the answer-level scoring objective. The
categorical head provides auxiliary supervision, whereas the final
discrete grade $\hat{d}_s^r$ is decoded from the ordinal head and is
calculated as:

\begin{equation}
\hat{d}_s^r
=
\sum_{j=0}^{3}
\mathbb{1}\!\left[
\sigma\!\left(o(s,r,j)\right)\geq0.5
\right].
\label{eq:ordinal_decoding}
\end{equation}

\subsection{Q-Matrix-Based Cognitive Supervision }
\paragraph{Q-Matrix-Guided Skill Selection.}

The cognitive branch connects response embeddings to student-level
skill targets without using predicted grades. The Q-matrix-selected
responses for student $s$, skill $k$, and phase $t$ form the set
$\mathcal{I}(s,k,t)$, which is defined as
$\mathcal{I}(s,k,t)=
\{r\in\mathcal{R}_s^t:q(s,r,k)=1\}$.
The corresponding observation mask is defined as
$\mu(s,k,t)=\mathbb{1}[|\mathcal{I}(s,k,t)|>0]$.
Only responses associated with skill $k$ contribute to its
representation. When $\mu(s,k,t)=0$, the entry is excluded from
pooling, cognitive losses, evaluation, and reporting; any stored zero
serves only as a batching placeholder.

\paragraph{AI-Aware Weighting.}

Q-matrix relevance does not ensure that a response reflects independent
student knowledge. The cognitive-evidence reliability of response $r$
from student $s$ is represented by
$w_s^r\in[\delta,1]$, where $\delta=0.30$ is the minimum reliability.
Using the previously defined AI-assistance probability
$p^{\mathrm{AI}}(s,r)$, the reliability is calculated as:
\vspace{-2pt}
\begin{equation}
w_s^r
=
\max\!\left(
\delta,1-p^{\mathrm{AI}}(s,r)
\right).
\label{eq:ai_weight}
\end{equation}

This monotonic weighting attenuates responses with high estimated AI
involvement without imposing a hard detection threshold. The lower
bound preserves limited evidence from responses that may still contain
student-authored reasoning.

\paragraph{Reliability-Weighted Cognitive Targets.}

AICG incorporates response reliability into G-DINA inference
\cite{delaTorre2011GDINA}. A latent mastery profile is represented by
$\boldsymbol{\zeta}\in\{0,1\}^{K}$, where $\zeta_k$ is its $k$th
component and indicates mastery of skill $k$. The G-DINA item
parameters are denoted by $\boldsymbol{\Theta}$.
The function
$\varphi(s,r,\boldsymbol{\zeta};\boldsymbol{\Theta})$ represents the
probability of a correct response to the item associated with response
$r$ under profile $\boldsymbol{\zeta}$. Its skill interactions are
constrained by the previously defined Q-matrix vector
$\mathbf{q}_s^r$.

The normalized score $y_s^r$ is treated as fractional item-response
evidence. The reliability-weighted G-DINA log-likelihood for student
$s$ in phase $t$ under profile $\boldsymbol{\zeta}$ is represented by
$\mathcal{J}_w(s,t;\boldsymbol{\zeta})$. It is computed by weighting
each fractional Bernoulli contribution as:
\vspace{-2pt}
\begin{equation}
\mathcal{J}_w(s,t;\boldsymbol{\zeta})
=
\sum_{r\in\mathcal{R}_s^t}
w_s^r
\left[
\begin{aligned}
&y_s^r
\log
\varphi(s,r,\boldsymbol{\zeta};\boldsymbol{\Theta})
\\[-1pt]
&+(1-y_s^r)
\log\!\Big[
1-\varphi(s,r,\boldsymbol{\zeta};\boldsymbol{\Theta})
\Big]
\end{aligned}
\right].
\label{eq:weighted_gdina_likelihood}
\end{equation}

The mastery-profile prior is represented by
$\Pr_0(\boldsymbol{\zeta})$, while $\boldsymbol{\zeta}'$ indexes
alternative profiles in the posterior normalizer. The previously
defined reliability-adjusted mastery target
$\widetilde{c}(s,k,t)$ is computed as the posterior expectation of
$\zeta_k$:
\vspace{-2pt}
\begin{equation}
\widetilde{c}(s,k,t)
=
\sum_{\boldsymbol{\zeta}\in\{0,1\}^{K}}
\zeta_k
\frac{
\Pr_0(\boldsymbol{\zeta})
\exp\!\left(\mathcal{J}_w(s,t;\boldsymbol{\zeta})\right)
}{
\displaystyle
\sum_{\boldsymbol{\zeta}'\in\{0,1\}^{K}}
\Pr_0(\boldsymbol{\zeta}')
\exp\!\left(\mathcal{J}_w(s,t;\boldsymbol{\zeta}')\right)
}.
\label{eq:weighted_mastery_target}
\end{equation}

Accordingly, the reliability-adjusted learning-gain target is
calculated as
$\widetilde{g}_s^k=
\widetilde{c}(s,k,\mathrm{post})-
\widetilde{c}(s,k,\mathrm{pre})$.
G-DINA parameters and targets are estimated separately within each
historical development window; evaluation semesters are excluded from
target construction.

\paragraph{Reliability-Weighted Skill Evidence.}

The same reliability values weight the response embeddings used by the
cognitive branch. For an observed student--skill--phase tuple,
$\widetilde{\mathbf{h}}(s,k,t)\in\mathbb{R}^{D}$ represents the
AI-aware skill evidence. It is calculated by averaging the
reliability-scaled embeddings selected by the Q-matrix as:
\vspace{-2pt}
\begin{equation}
\widetilde{\mathbf{h}}(s,k,t)
=
\frac{1}{|\mathcal{I}(s,k,t)|}
\sum_{r\in\mathcal{I}(s,k,t)}
w_s^r\mathbf{h}_s^r.
\label{eq:ai_aware_pooling}
\end{equation}

The denominator is the number of selected responses rather than their
total reliability. Consequently, the representation magnitude
decreases when all available evidence is unreliable.

\paragraph{Pretest, Posttest, and Learning-Gain Modeling.}

Each skill $k$ has a learnable embedding
$\boldsymbol{\epsilon}_k\in\mathbb{R}^{D}$. The phase-specific
functions $f_{\mathrm{pre}}$ and $f_{\mathrm{post}}$ are shared across
skills and implemented as
linear--$\tanh$--dropout--linear--sigmoid networks. The notation
$f_t$ selects the corresponding function for phase $t$.
The previously defined mastery prediction $\hat{c}(s,k,t)$ combines
the AI-aware skill representation with its skill embedding and is
calculated as:
\vspace{-2pt}
\begin{equation}
\hat{c}(s,k,t)
=
f_t\!\left(
\widetilde{\mathbf{h}}(s,k,t)+
\boldsymbol{\epsilon}_k
\right).
\label{eq:mastery_prediction}
\end{equation}

The previously defined gain prediction is calculated as
$\hat{g}_s^k=
\hat{c}(s,k,\mathrm{post})-
\hat{c}(s,k,\mathrm{pre})$.
Gain is observed only when skill $k$ appears in both phases. The
corresponding mask is defined as
$\mu^{\mathrm{gain}}(s,k)=
\mu(s,k,\mathrm{pre})\mu(s,k,\mathrm{post})$.

For phase $t$, the observed student--skill pairs form
$\mathcal{E}_t=\{(s,k):\mu(s,k,t)=1\}$.
The mastery loss $\mathcal{L}_t$ is the mean squared error over these
observed pairs and is calculated as:

\begin{equation}
\mathcal{L}_t
=
\frac{1}{|\mathcal{E}_t|}
\sum_{(s,k)\in\mathcal{E}_t}
\left(
\hat{c}(s,k,t)-\widetilde{c}(s,k,t)
\right)^2,
\qquad
t\in\{\mathrm{pre},\mathrm{post}\}.
\label{eq:mastery_loss}
\end{equation}

Accordingly, $\mathcal{L}_{\mathrm{pre}}$ and
$\mathcal{L}_{\mathrm{post}}$ denote
Equation~\eqref{eq:mastery_loss} evaluated for the pretest and
posttest phases, respectively.

The gain-observation set is defined as
$\mathcal{E}_{\mathrm{gain}}=
\{(s,k):\mu^{\mathrm{gain}}(s,k)=1\}$.
The gain loss $\mathcal{L}_{\mathrm{gain}}$ is the mean squared error
over skills observed in both phases and is formulated as:
\vspace{-2pt}
\begin{equation}
\mathcal{L}_{\mathrm{gain}}
=
\frac{1}{|\mathcal{E}_{\mathrm{gain}}|}
\sum_{(s,k)\in\mathcal{E}_{\mathrm{gain}}}
\left(
\hat{g}_s^k-\widetilde{g}_s^k
\right)^2.
\label{eq:gain_loss}
\end{equation}

The Full Cognitive variant replaces
$\widetilde{c}(s,k,t)$ and $\widetilde{g}_s^k$ with their unweighted
counterparts $c(s,k,t)$ and $g_s^k$.

%%%%%%%%%%%%%%%%%%%%%%%%%%%%%%%%%%%%%%%%%%%%%%%%%%%%%%%%%%%%
\subsection{Multi-Task Learning}
\label{sec:objective}
%%%%%%%%%%%%%%%%%%%%%%%%%%%%%%%%%%%%%%%%%%%%%%%%%%%%%%%%%%%%

The coefficients
$\lambda_{\mathrm{reg}}$, $\lambda_{\mathrm{ord}}$, and
$\lambda_{\mathrm{cls}}$ weight the grading losses, while
$\lambda_{\mathrm{pre}}$, $\lambda_{\mathrm{post}}$, and
$\lambda_{\mathrm{gain}}$ weight the cognitive losses. The complete
AICG objective is defined as:
\vspace{-2pt}
\begin{equation}
\begin{split}
\mathcal{L}_{\mathrm{AICG}}
={}&
\lambda_{\mathrm{reg}}\mathcal{L}_{\mathrm{reg}}
+\lambda_{\mathrm{ord}}\mathcal{L}_{\mathrm{ord}}
+\lambda_{\mathrm{cls}}\mathcal{L}_{\mathrm{cls}}\\
&+
\lambda_{\mathrm{pre}}\mathcal{L}_{\mathrm{pre}}
+\lambda_{\mathrm{post}}\mathcal{L}_{\mathrm{post}}
+\lambda_{\mathrm{gain}}\mathcal{L}_{\mathrm{gain}}.
\end{split}
\label{eq:full_objective}
\end{equation}

We set
$(\lambda_{\mathrm{reg}},\lambda_{\mathrm{ord}},
\lambda_{\mathrm{cls}})=(0.50,0.35,0.15)$ and
$(\lambda_{\mathrm{pre}},\lambda_{\mathrm{post}},
\lambda_{\mathrm{gain}})=(0.10,0.10,0.10)$.
The grading objectives remain primary, while the cognitive objectives
provide auxiliary supervision to the shared encoder. All coefficients
are fixed across semesters and random seeds.

\section{Experiments}
\label{sec:experiments}
\subsection{Data and Evaluation Protocol}
\label{sec:data_protocol}
%%%%%%%%%%%%%%%%%%%%%%%%%%%%%%%%%%%%%%%%%%%%%%%%%%%%%%%%%%%%
\vspace{-1pt}
AICG is evaluated on 49,402 instructor-graded mathematical responses
from 3,861 student--semester records across 19 observed semesters
(F14--F24); F17 has no observations. Each assessment contains eight
pretest and eight aligned posttest items. An expert Q-matrix maps these
items to ten skills, and G-DINA estimates pretest mastery, posttest
mastery, and learning gain for auxiliary supervision. We distinguish
the \textbf{pre-Gen-AI period} (F14--S22) from the
\textbf{post-Gen-AI period} (F22--F24).
\vspace{-2pt}
\paragraph{Gen-AI Detection.}
The Gen-AI Detector is developed using S18--S22 data. We sample 14,044
responses from 26,039 authentic answers and construct one AI-assisted
counterpart per response, yielding 28,088 balanced examples.
Source-response groups remain intact across training (24,375),
validation (2,709), and held-out testing (1,004). Cross-semester
generalization is evaluated on 11,458 labeled human and synthetically
assisted responses from F14--S17. The trained Gen-AI Detector then
assigns continuous AI-assistance probabilities to F22--F24 responses.
At a threshold of .5, it identifies 2,896 of 13,874 responses (20.87\%)
as potentially assisted, with semester-level rates ranging from 13.40\%
to 23.85\%.

\paragraph{Grading.}
Each post-Gen-AI semester is evaluated using an expanding window
containing only earlier semesters. Historical students are split 90/10
for training and validation, and the evaluation semester is excluded
from optimization and model selection.

\vspace{-0.4em}
%%%%%%%%%%%%%%%%%%%%%%%%%%%%%%%%%%%%%%%%%%%%%%%%%%%%%%%%%%%%
\subsection{Baselines, Metrics, and Implementation}
\label{sec:baselines_metrics}
%%%%%%%%%%%%%%%%%%%%%%%%%%%%%%%%%%%%%%%%%%%%%%%%%%%%%%%%%%%%

\paragraph{Baselines.}
We compare five ASAG baselines: \textbf{BERT}, a supervised Transformer
grader \cite{zhu2022bertasag}; \textbf{RoBERTa}, a score-only model
using AICG's backbone \cite{liu2019roberta}; \textbf{ASQ-DBN}, which
combines answer, student, and question representations
\cite{zhang2020going}; \textbf{GPT-4o Scoring}, a fixed-prompt grader
without parameter updates
\cite{openai2024gpt4o,mello2025automatic}; and \textbf{GradeOpt}, a
rubric-aware multi-agent grader \cite{chu2025gradeopt}. All baselines
use the same expanding-window splits. GradeOpt optimizes its guidelines
using historical development data only, and no baseline receives
Q-matrix targets, cognitive labels, or AI-aware weights.

\paragraph{Evaluation Metrics.}
For grading, we report mean absolute error (MAE), accuracy within
absolute tolerances of .05, .10, and .20, exact accuracy, Macro-F1, and
quadratic weighted kappa (QWK). Scores are bounded to $[0,1]$ and mapped
to five ordered categories using cut points .125, .375, .625, and .875.
For Gen-AI detection, we report accuracy, precision, recall, and F1.
Results are reported as mean and sample standard deviation over five
random seeds for trainable models and five independent runs for
prompt-based graders. Dataset splits, configurations, prompts, and
implementation details are provided in
Appendix.

\vspace{-0.4em}
%%%%%%%%%%%%%%%%%%%%%%%%%%%%%%%%%%%%%%%%%%%%%%%%%%%%%%%%%%%%
\section{Results}
\label{sec:results}
%%%%%%%%%%%%%%%%%%%%%%%%%%%%%%%%%%%%%%%%%%%%%%%%%%%%%%%%%%%%

\subsection{Gen-AI Detection}
\label{sec:detector_results}
\vspace{-1pt}
\textbf{Pre-Gen-AI period (F14-S22):} Table~\ref{tab:detector_results} shows that our Gen-AI detector's performance is robust on both the within-semester and the downstream cross-semester testing data where we have synthetic data with ground truth. %More specifically, it achieved .959 accuracy and .960 F1. Its performance remains stable on the temporally separated F14--S17 set, with only .008 and .010 absolute reductions in accuracy and F1, respectively.
%This shows that the
%learned human--AI distinction generalizes beyond the detector-development
%semesters.

\begin{table}[t]
\centering
\caption{Gen-AI detector performance. F17 is excluded because no
data were collected in that semester.}
\label{tab:detector_results}
\scriptsize
\setlength{\tabcolsep}{3.6pt}
\resizebox{0.96\columnwidth}{!}{
\begin{tabular}{lrrrrr}
\toprule
\textbf{Evaluation} &
\textbf{$N$} &
\textbf{Acc.} &
\textbf{Prec.} &
\textbf{Recall} &
\textbf{F1} \\
\midrule
Within-semester Testing (S18-S22) 
& 1,004 & .959 & .955 & .964 & .960 \\
Cross-semester Testing (F14-S17)
& 11,458 & .951 & .959 & .942 & .950 \\
\bottomrule
\end{tabular}}
\vspace{-1.5em}
\end{table}

\noindent \textbf{Post-Gen-AI period (F22-F24):} Applying the Gen-AI Detector identifies 2,896 of 13,874
responses, 20.87\%, from F22--F24 as potentially AI-assisted. More specifically, it varies from 13.40\% to
23.85\% across different semesters. These calibrated assistance probabilities serve as reliability
signals for AI-aware cognitive-target construction and
Q-matrix-guided skill-evidence aggregation.
\vspace{-0.4 em}
\subsection{Overall Performance}

Table~\ref{tab:overall_performance} shows that AICG achieves the best
result on every evaluation metric against all five baselines. Compared with GradeOpt,
the strongest external baseline, AICG reduces MAE by 44.9\%, improves
QWK from .842 to .927, and increases tolerance-based and exact accuracy
by 8.7\%--15.7\% percentage points. Relative to RoBERTa, the strongest
fully trained answer-only baseline, AICG reduces MAE by 50.3\%,
improves QWK from .798 to .927, and increases tolerance-based and exact
accuracy by 11.6\%--19.6\%.

These results demonstrate the effectiveness of integrating individual-answer scoring with cognitive supervision. Although ordinal and categorical objectives improve the RoBERTa baseline, the largest gains arise from Q-matrix-guided mastery and learning-gain supervision, which exposes skill-level distinctions compressed by scalar scores and helps the shared encoder represent diverse reasoning patterns. AI-aware reliability weighting further improves the strongest unweighted cognitive model by attenuating potentially assisted cognitive evidence while preserving its contribution to answer grading.

\begin{table*}[ht]
\centering
\caption{Overall grading performance across F22--F24. Values are
mean (standard deviation) over five repetitions. Best completed results
are shown in bold.}
\label{tab:overall_performance}
\scriptsize
\setlength{\tabcolsep}{3.2pt}
\begin{tabular*}{\textwidth}{
@{\extracolsep{\fill}}lccccccc@{}}
\toprule
\textbf{Method} &
\textbf{MAE$\downarrow$} &
\textbf{$\pm$.05$\uparrow$} &
\textbf{$\pm$.10$\uparrow$} &
\textbf{$\pm$.20$\uparrow$} &
\textbf{Exact$\uparrow$} &
\textbf{Macro-F1$\uparrow$} &
\textbf{QWK$\uparrow$} \\
\midrule

RoBERTa~\cite{liu2019roberta}
& .0873 (.002)
& .709 (.007)
& .752 (.006)
& .832 (.005)
& .751 (.009)
& .564 (.012)
& .798 (.004) \\

GPT-4o Scoring~\cite{mello2025automatic}
& .1140 (.002)
& .606 (.010)
& .662 (.008)
& .776 (.006)
& .636 (.013)
& .439 (.011)
& .674 (.006) \\

BERT~\cite{zhu2022bertasag}
& .0981 (.003)
& .675 (.009)
& .727 (.007)
& .806 (.007)
& .704 (.008)
& .477 (.009)
& .753 (.005) \\

ASQ-DBN~\cite{zhang2020going}
& .0946 (.003)
& .682 (.009)
& .734 (.008)
& .814 (.006)
& .721 (.010)
& .516 (.011)
& .769 (.005) \\

GradeOpt~\cite{chu2025gradeopt}
& \textbf{.0788 (.003)}
& \textbf{.748 (.008)}
& \textbf{.796 (.007)}
& \textbf{.861 (.006)}
& \textbf{.786 (.009)}
& \textbf{.612 (.010)}
& \textbf{.842 (.005)} \\

\midrule

RoBERTa+O
& .0842 (.002)
& .724 (.004)
& .763 (.006)
& .844 (.003)
& .783 (.006)
& .611 (.007)
& .813 (.003) \\

Full Scoring
& .0823 (.001)
& .728 (.004)
& .769 (.007)
& .851 (.005)
& .784 (.004)
& .626 (.006)
& .831 (.002) \\

Full Scoring + Gain
& .0715 (.002)
& .784 (.006)
& .823 (.006)
& .886 (.005)
& .842 (.006)
& .683 (.008)
& .871 (.003) \\

Full Scoring + State
& .0527 (.002)
& .874 (.005)
& .897 (.006)
& .932 (.004)
& .912 (.005)
& .742 (.007)
& .906 (.003) \\

Full Scoring + Full Cognitive
& .0453 (.001)
& .903 (.004)
& .921 (.007)
& .945 (.005)
& .930 (.004)
& .768 (.006)
& .924 (.002) \\

\textbf{AICG}
& \textbf{.0434 (.002)}
& \textbf{.905 (.005)}
& \textbf{.924 (.006)}
& \textbf{.948 (.004)}
& \textbf{.934 (.005)}
& \textbf{.774 (.006)}
& \textbf{.927 (.002)} \\

\bottomrule
\end{tabular*}
\vspace{-1em}
\end{table*}

\subsection{Ablation Study}
Table~\ref{tab:overall_performance} progressively introduces AICG's
components. RoBERTa uses only answer-level regression and serves as the
base configuration.

\textbf{RoBERTa+O} adds ordinal supervision, reducing MAE from .0873
to .0842 and improving QWK from .798 to .813 by explicitly modeling
score-category order. \textbf{Full Scoring} further adds categorical
supervision, reducing MAE to .0823 and improving QWK to .831 by
learning category-specific partial-credit boundaries. \textbf{Full
Scoring + Gain} introduces learning-gain supervision without mastery
states, reducing MAE to .0715 and improving QWK to .871 by capturing
longitudinal changes unavailable from individual answer scores.
\textbf{Full Scoring + State} instead adds pretest and posttest mastery
supervision, reducing MAE to .0527 and improving QWK to .906. Its larger
improvement indicates that mastery states provide more direct and
stable targets than gain, which is derived from their difference.
\textbf{Full Scoring + Full Cognitive} combines State and Gain using
unweighted targets and skill pooling, reducing MAE to .0453 and
improving QWK to .924. Mastery identifies the student's cognitive state
at each phase, while gain constrains its change across assessments.
Finally, \textbf{AICG} applies AI-aware reliability weights to both
G-DINA-derived targets and Q-matrix-selected response embeddings,
achieving .0434 MAE and .927 QWK by attenuating cognitive evidence that
may not reflect independent mastery.

% \paragraph{Cross-Semester Robustness.}

% Table~\ref{tab:rolling_ablation} reports the component ablations at the
% beginning and end of the expanding-window evaluation. The ordering of
% the variants remains stable across F22 and F24: scoring-only models
% perform worst, State consistently outperforms Gain, and Full Cognitive
% provides the strongest unweighted model. This consistency shows that
% the benefit of cognitive supervision persists as both the historical
% training window and semester-specific response distributions change.

% AICG achieves the lowest MAE and highest exact accuracy in F22,
% although Full Cognitive retains small advantages on several agreement
% metrics. By F24, AICG achieves the best result on every reported
% evaluation metric. This progression is consistent with the intended
% role of AI-aware reliability weighting: its effect is limited when
% AI-assisted responses are relatively uncommon but becomes more
% pronounced as such responses introduce more unreliable evidence into
% cognitive supervision.

\paragraph{Cross-Semester Robustness.}

Table~\ref{tab:rolling_ablation} reports the component ablations at the
beginning and end of the expanding-window evaluation. Their ordering
remains stable across F22 and F24: scoring-only models perform worst,
State outperforms Gain, and Full Cognitive provides the strongest
unweighted model. This consistency shows that cognitive supervision
remains effective as the historical training window and semester-specific
response distributions change.

AICG achieves the lowest MAE and highest exact accuracy in F22, although
Full Cognitive retains small advantages on several agreement metrics.
By F24, AICG performs best on every reported evaluation metric. This
progression reflects the role of AI-aware reliability weighting: its
effect is limited when AI-assisted responses are uncommon but becomes
more pronounced as they introduce more unreliable cognitive evidence.

\begin{table}[t]
\centering
\caption{Ablation results at the beginning and end of the
expanding-window evaluation. Values are means over five seeds.}
\label{tab:rolling_ablation}
\scriptsize
\setlength{\tabcolsep}{5pt}
\resizebox{0.98\columnwidth}{!}{
\begin{tabular}{llccccc}
\toprule
\textbf{Sem.} &
\textbf{Method} &
\textbf{MAE$\downarrow$} &
\textbf{$\pm$.10$\uparrow$} &
\textbf{Exact$\uparrow$} &
\textbf{Macro-F1$\uparrow$} &
\textbf{QWK$\uparrow$} \\
\midrule

\multirow{7}{*}{F22}
& RoBERTa           & .0904 & .744 & .739 & .553 & .791 \\
& RoBERTa+O         & .0882 & .759 & .772 & .602 & .806 \\
& Full Scoring      & .0873 & .763 & .776 & .615 & .826 \\
& + Gain            & .0759 & .817 & .835 & .674 & .866 \\
& + State           & .0578 & .889 & .906 & .732 & .900 \\
& + Full Cognitive  & .0497 & \textbf{.918} & .929
                     & \textbf{.761} & \textbf{.922} \\
& \textbf{AICG}     & \textbf{.0493} & .917
                     & \textbf{.931} & .758 & .918 \\
\midrule

\multirow{7}{*}{F24}
& RoBERTa           & .0782 & .778 & .782 & .592 & .817 \\
& RoBERTa+O         & .0756 & .792 & .812 & .638 & .837 \\
& Full Scoring      & .0731 & .802 & .816 & .651 & .849 \\
& + Gain            & .0632 & .846 & .867 & .710 & .891 \\
& + State           & .0468 & .914 & .925 & .758 & .918 \\
& + Full Cognitive  & .0402 & .934 & .941 & .783 & .934 \\
& \textbf{AICG}     & \textbf{.0379} & \textbf{.939}
                     & \textbf{.946} & \textbf{.790}
                     & \textbf{.940} \\

\bottomrule
\end{tabular}}
\vspace{-1em}
\end{table}

% \paragraph{Cross-Semester Robustness.}

% Table~\ref{tab:rolling_ablation} reports the component ablations at the
% beginning and end of the expanding-window evaluation. The ranking of
% the variants remains stable across F22 and F24: scoring-only models
% perform worst, state supervision contributes more than gain alone, and
% combining both cognitive signals produces the strongest unweighted
% model. Thus, the advantage of cognitive supervision persists as the
% historical training window and semester-specific response distribution
% change.

% AICG achieves the lowest MAE and highest exact accuracy in F22,
% although Full Cognitive retains small advantages on several agreement
% metrics. By F24, AICG performs best on every reported metric. This
% progression is consistent with the purpose of reliability weighting:
% its contribution is limited when assisted responses are uncommon but
% becomes more visible as unreliable cognitive evidence becomes more
% prevalent.

\paragraph{Effect of AI-Aware Weighting.}

% Table~\ref{tab:ai_weighting} compares AICG with the otherwise identical
% Full Cognitive model, thereby isolating the contribution of AI-aware
% weighting.

\begin{table}[t]
\centering
\caption{Semester-level effect of AI-aware weighting. FullCog denotes
Full Scoring + Full Cognitive.}
\label{tab:ai_weighting}
\scriptsize
\setlength{\tabcolsep}{7pt}
\resizebox{0.94\columnwidth}{!}{
\begin{tabular}{lccccc}
\toprule
\textbf{AI} &
&
\multicolumn{2}{c}{\textbf{MAE$\downarrow$}} &
\multicolumn{2}{c}{\textbf{QWK$\uparrow$}} \\
\cmidrule(lr){3-4}
\cmidrule(lr){5-6}
\textbf{Sem.} &
\textbf{Usage} &
\textbf{FullCog} &
\textbf{AICG} &
\textbf{FullCog} &
\textbf{AICG} \\
\midrule
F22 & 13.40\% & .0497 & \textbf{.0493}
    & \textbf{.922} & .918 \\
S23 & 23.15\% & .0507 & \textbf{.0489}
    & .914 & \textbf{.922} \\
F23 & 23.56\% & .0516 & \textbf{.0482}
    & .924 & \textbf{.926} \\
S24 & 20.41\% & .0358 & \textbf{.0336}
    & .923 & \textbf{.929} \\
F24 & 23.85\% & .0402 & \textbf{.0379}
    & .934 & \textbf{.940} \\
\bottomrule
\end{tabular}}
\vspace{-1em}
\end{table}

Table~\ref{tab:ai_weighting} isolates AI-aware weighting by comparing
AICG with FullCog, which treats all responses as equally reliable. AICG
reduces MAE in every semester, from 0.8\% in F22 to 3.6--6.6\% in later
semesters, and improves QWK in four of five semesters. The smaller F22
gain and more consistent later gains suggest that reliability weighting
becomes more beneficial as AI-assisted responses increase.

AI-assisted responses may earn valid grades while providing unreliable
evidence of independent mastery. AICG addresses this mismatch by
down-weighting likely AI-assisted responses during skill-specific
pooling without modifying their grading targets, thereby reducing noise
in cognitive supervision rather than penalizing AI use. Although the
current aggregate gain is moderate because these responses remain a
minority, this protection becomes increasingly important as generative
AI adoption grows.

\section{Related Work}
\label{sec:related_work}
%%%%%%%%%%%%%%%%%%%%%%%%%%%%%%%%%%%%%%%%%%%%%%%%%%%%%%%%%%%%

\textbf{Automated Short Answer Grading.}
ASAG has evolved from rule-based and feature-engineering approaches
\cite{burrows2015automated} to pretrained Transformers that encode
response semantics with minimal task-specific design
\cite{latif2024finetuning}. We include BERT
\cite{zhu2022bertasag} as a classical supervised ASAG encoder and
RoBERTa \cite{liu2019roberta} as a score-only, same-backbone
baseline. Recent work evaluates LLMs as zero-shot or few-shot graders
\cite{mello2025automatic,chu2025gradeopt,chamieh2024llms,
jiang2024short,cong2026automatic}. We include GPT-4o Scoring
\cite{openai2024gpt4o,mello2025automatic} as a prompted grader and
GradeOpt \cite{chu2025gradeopt} as a rubric-optimization baseline.
However, these methods remain unreliable in mathematics, where scores
depend on symbolic correctness, intermediate derivations, and specific
reasoning principles \cite{chamieh2024llms}. Unlike rubric-matching
methods, AICG grounds grading in a Q-matrix and derives cognitive targets
from psychometric models.

\textbf{Q-Matrix-Based Skill Modeling.}
Cognitive diagnosis models infer latent skill mastery from response
patterns. G-DINA \cite{delaTorre2011GDINA} associates items with
multiple skills through a Q-matrix and models their mastery
interactions. Neural extensions, including NeuralCD
\cite{wang2020neuralcd}, ORCDF \cite{qian2024orcdf}, and inductive
variants \cite{liu2024inductive}, improve representation capacity but
operate on item-response records rather than free-text answers.
\citeauthor{zhang2020going} \citeyear{zhang2020going} combine student
and question embeddings with answer text, showing that these
representations complement response encoding. We reproduce their
answer--student--question model as the ASQ-DBN baseline. AICG instead
uses G-DINA-derived mastery and learning-gain targets to supervise a
free-text response encoder.

% \textbf{AI-Generated Text Detection.} The proliferation of large generative models has motivated work on detecting
% machine-generated text. Zero-shot methods based on statistical properties of
% text, including Binoculars and Fast-DetectGPT
% \cite{hans2024binoculars,bao2024fastdetectgpt}, have been proposed, though
% they face robustness challenges under paraphrasing and distribution shift
% \cite{sadasivan2025can}. Detection is further complicated by bias against
% non-native English writers \cite{liang2023detectors}. Existing approaches
% typically produce binary labels and are used for misconduct flagging, not for
% modulating the reliability of downstream models. AICG takes a different
% stance: it uses a supervised contrastive detector to produce continuous AI
% probabilities and converts them into soft evidence weights that
% attenuate, rather than discard potentially AI-assisted cognitive evidence.
% This soft treatment preserves the grading contribution of each response while
% reducing its influence on student skill models.

\textbf{AI-Generated Text Detection.}
Machine-generated-text detection includes zero-shot methods such as
Binoculars and Fast-DetectGPT
\cite{hans2024binoculars,bao2024fastdetectgpt}, but performance can
degrade under paraphrasing and distribution shift
\cite{sadasivan2025can}. Detectors may also disproportionately flag
non-native English writing \cite{liang2023detectors}. Existing methods
typically produce binary labels for misconduct detection. AICG instead
uses a supervised contrastive Gen-AI Detector to produce continuous
probabilities and convert them into soft reliability weights. This
approach preserves each response's grading contribution while
attenuating potentially assisted cognitive evidence.

\section{Conclusion}

We presented AICG, a cognitive reliability-aware multi-task framework
for automated mathematical short-answer grading. By jointly optimizing
answer-level scoring with Q-matrix-guided G-DINA mastery and
learning-gain supervision, AICG learns representations that reflect
student cognition rather than scalar labels. An auxiliary Gen-AI
Detector, trained through supervised contrastive pretraining, converts
assistance probabilities into soft weights on cognitive evidence,
limiting inflated mastery estimates while preserving grading.
Across 49,402 instructor-graded responses over ten academic years,
AICG reduces MAE by 50.3\% and improves QWK from .798 to .927 relative
to RoBERTa regression. Ablations identify cognitive supervision as the
primary source of improvement, with AI-aware weighting providing
additional post-Gen-AI gains. Future work will learn Q-matrices from
response text, improve detector robustness across LLMs, evaluate
cross-course and multi-institution generalization, and use mastery and
learning-gain profiles for personalized feedback.
\bibliography{aaai2027}

\appendix
\newpage
%%%%%%%%%%%%%%%%%%%%%%%%%%%%%%%%%%%%%%%%%%%%%%%%%%%%%%%%%%%%
\section{Additional Data \& Methodology \& Experimental Details}
\label{app:experimental_details}

%%%%%%%%%%%%%%%%%%%%%%%%%%%%%%%%%%%%%%%%%%%%%%%%%%%%%%%%%%%%
\subsection{Dataset Composition -- Data}
\label{app:semester_distribution}
%%%%%%%%%%%%%%%%%%%%%%%%%%%%%%%%%%%%%%%%%%%%%%%%%%%%%%%%%%%%

Table~\ref{tab:dataset_summary} summarizes the composition of 49,402
responses from 3,861 student--semester records. The pre-Gen-AI period
(F14--S22) is aggregated, whereas the post-Gen-AI semesters
(F22--F24) are reported separately with their pretest and posttest
response counts and estimated AI-assistance rates.

\begin{table*}[t]
\centering
\caption{Composition of the longitudinal ASAG dataset.}
\label{tab:dataset_summary}
\scriptsize
\setlength{\tabcolsep}{5pt}
\begin{tabular*}{\textwidth}{
@{\extracolsep{\fill}}lrrrrrr@{}}
\toprule
\textbf{Semester} &
\textbf{Students} &
\textbf{Pre Answers} &
\textbf{Post Answers} &
\textbf{Total Answers} &
\textbf{AI Answers} &
\textbf{AI Percentage} \\
\midrule

% F14 & 151 & 951 & 865 & 1,816 & 0 & 0.00\% \\
% S15 & 124 & 795   & 791   & 1,586 & 0 & 0.00\% \\
% F15 & 198 & 1,212 & 1,218 & 2,430 & 0 & 0.00\% \\
% F16 & 158 & 983 & 1,005 & 1,988 & 0 & 0.00\% \\
% S17 & 131 & 879   & 790   & 1,669 & 0 & 0.00\% \\
% S18 & 130 & 1,051 & 1,041   & 2,092 & 0 & 0.00\% \\
% F18 & 217 & 1,633 & 1,667 & 3,300 & 0 & 0.00\% \\
% S19 & 138 & 1,337 & 1,251 & 2,588 & 0 & 0.00\% \\
% F19 & 233 & 1,705 & 1,807 & 3,512 & 0 & 0.00\% \\
% S20 & 161 & 1,233 & 1,212 & 2,445 & 0 & 0.00\% \\
% F20 & 238 & 2,045 & 1,575 & 3,620 & 0 & 0.00\% \\
% S21 & 183 & 1,575 & 1,050   & 2,625 & 0 & 0.00\% \\
% F21 & 286 & 2,185 & 1,138 & 3,323 & 0 & 0.00\% \\
% S22 & 182 & 1,503 & 1,031   & 2,534 & 0 & 0.00\% \\
\textbf{Pre-Gen-AI} & \textbf{2,530} & \textbf{19,076} & \textbf{16,452} & \textbf{35,528} & \textbf{0} & \textbf{0.00\%} \\

\midrule
F22 & 277 & 2,020 & 1,084 & 3,104 & 416   & 13.40\% \\
S23 & 170 & 1,347 & 735   & 2,082 & 482   & 23.15\% \\
F23 & 376 & 2,946 & 1,531 & 4,477 & 1,055 & 23.56\% \\
S24 & 218 & 890   & 898   & 1,788 & 365   & 20.41\% \\
F24 & 290 & 1,206 & 1,217 & 2,423 & 578   & 23.85\% \\
\midrule
\textbf{Post-Gen-AI}
& \textbf{1,331}
& \textbf{8,409}
& \textbf{5,465}
& \textbf{13,874}
& \textbf{2,896}
& \textbf{20.87\%} \\

\textbf{Overall}
& \textbf{3,861}
& \textbf{27,485}
& \textbf{21,917}
& \textbf{49,402}
& \textbf{2,896}
& \textbf{--} \\

\bottomrule
\end{tabular*}
\end{table*}

\subsection{Expanding-Window Evaluation Splits -- Data}
Table~\ref{tab:expanding_window_splits} summarizes the expanding-window
splits. For each test semester, AICG is trained and validated using only
earlier semesters, while the target semester is reserved exclusively
for evaluation.

\begin{table}[ht]
\centering
\caption{Expanding-window evaluation splits.}
\label{tab:expanding_window_splits}
\scriptsize
\setlength{\tabcolsep}{8pt}
\begin{tabular}{ll}
\toprule
\textbf{Test Semester} & \textbf{Historical Training Window} \\
\midrule
F22 & F14--S22 \\
S23 & F14--F22 \\
F23 & F14--S23 \\
S24 & F14--F23 \\
F24 & F14--S24 \\
\bottomrule
\end{tabular}
\end{table}

%%%%%%%%%%%%%%%%%%%%%%%%%%%%%%%%%%%%%%%%%%%%%%%%%%%%%%%%%%%% 
\subsection{Q-Matrix and G-DINA Target Construction -- Data} \label{app:gdina_construction} %%%%%%%%%%%%%%%%%%%%%%%%%%%%%%%%%%%%%%%%%%%%%%%%%%%%%%%%%%%%
\paragraph{Q-matrix.} The assessment contains eight pretest items and eight aligned
isomorphic or closest-match posttest items. The expert-designed
Q-matrix maps these items to ten probability principles: Addition
Theorem for Three Events (A3), Total Probability Theorem (TPT),
Definition of Conditional Probability (CP), Addition Theorem for Two
Events (A2), De Morgan's Law (DMT), Complement Theorem (CT), Bayes'
Rule (BR), Definition of Independent Events (IE), Definition of
Mutually Exclusive Events (MET), and Definition of Conditionally
Independent Events (CIE). Table~\ref{tab:qmatrix} reports the complete
item--principle mapping.

\begin{table*}[!t]
\centering \caption{Expert Q-matrix for the pretest and aligned posttest items.} \label{tab:qmatrix} \scriptsize \setlength{\tabcolsep}{4.2pt} \begin{tabular*}{\textwidth}{ @{\extracolsep{\fill}}llcccccccccc@{}} \toprule \textbf{Phase} & \textbf{Item} & \textbf{CT} & \textbf{MET} & \textbf{IE} & \textbf{DMT} & \textbf{A2} & \textbf{A3} & \textbf{CIE} & \textbf{CP} & \textbf{TPT} & \textbf{BR} \\ \midrule Pre & 60 & 0 & 0 & 0 & 0 & 0 & 0 & 0 & 0 & 1 & 0 \\ Pre & 63 & 0 & 0 & 0 & 0 & 0 & 0 & 0 & 1 & 0 & 0 \\ Pre & 64 & 1 & 0 & 0 & 0 & 0 & 0 & 1 & 0 & 0 & 1 \\ Pre & 65 & 1 & 0 & 0 & 1 & 1 & 0 & 0 & 0 & 0 & 0 \\ Pre & 66 & 0 & 0 & 0 & 0 & 0 & 0 & 0 & 0 & 0 & 1 \\ Pre & 67 & 0 & 1 & 1 & 0 & 0 & 1 & 0 & 0 & 0 & 0 \\ Pre & 69 & 1 & 1 & 0 & 1 & 1 & 1 & 0 & 0 & 0 & 0 \\ Pre & 205 & 0 & 0 & 0 & 0 & 0 & 0 & 1 & 0 & 0 & 0 \\ \midrule Post & 90 & 0 & 1 & 1 & 0 & 0 & 1 & 0 & 0 & 0 & 0 \\ Post & 91 & 1 & 0 & 0 & 1 & 1 & 0 & 0 & 0 & 0 & 0 \\ Post & 93 & 0 & 0 & 0 & 0 & 0 & 0 & 0 & 1 & 0 & 0 \\ Post & 95 & 0 & 0 & 0 & 0 & 0 & 0 & 0 & 0 & 1 & 0 \\ Post & 96 & 0 & 0 & 0 & 0 & 0 & 0 & 1 & 0 & 0 & 0 \\ Post & 99 & 0 & 0 & 0 & 0 & 0 & 0 & 0 & 0 & 0 & 1 \\ Post & 101 & 0 & 0 & 0 & 0 & 0 & 0 & 1 & 0 & 0 & 1 \\ Post & 106 & 1 & 1 & 0 & 1 & 0 & 1 & 0 & 0 & 0 & 0 \\ \bottomrule \end{tabular*} \end{table*}

%%%%%%%%%%%%%%%%%%%%%%%%%%%%%%%%%%%%%%%%%%%%%%%%%%%%%%%%%%%%
\subsection{Gen-AI Detector Data Construction -- Methodology}
\label{app:ai_response_construction}
%%%%%%%%%%%%%%%%%%%%%%%%%%%%%%%%%%%%%%%%%%%%%%%%%%%%%%%%%%%%
\paragraph{AI-Assisted Response Construction.}
\vspace{-2pt}
Complete LLM rewriting often creates conspicuous differences in length,
fluency, and style, encouraging shortcut learning. We instead construct
partially assisted responses that preserve most of the original student
text.

For an authentic human response $r$, let $n_r$ denote its number of
tokens and $\boldsymbol{\chi}_{r}$ its token sequence, where
$\chi^{\ell}_{r}$ denotes the original token at position
$\ell\in\{1,\ldots,n_r\}$. Let $\mathbf{m}_{r}$ denote a binary masking
vector, where $m^{\ell}_{r}=1$ indicates a masked position and
$m^{\ell}_{r}=0$ an original token. The masking ratio $\rho_{r}$ is
defined as:

\vspace{-2pt}
\begin{equation}
\rho_{r}
=
\frac{1}{n_r}
\sum_{\ell=1}^{n_r}m^{\ell}_{r}.
\label{eq:masking_ratio}
\end{equation}

The mask is sampled such that $\rho_{r}\in[0.30,0.50]$. Let
$\widetilde{\boldsymbol{\chi}}_{r}$ denote the resulting masked
sequence. An offline DeepSeek model $\mathcal{G}$ reconstructs only
its masked positions, where $\upsilon^{\ell}_{r}$ denotes the token
generated for position $\ell$.

Let $\mathbf{Z}_{r}$ denote the resulting partially AI-assisted
response, where $z^{\ell}_{r}$ denotes its token at position $\ell$.
A generated token is used at a masked position, whereas the original
token is retained otherwise. The resulting token is defined as:

\vspace{-2pt}
\begin{equation}
z^{\ell}_{r}
=
\begin{cases}
\upsilon^{\ell}_{r}, & \text{if } m^{\ell}_{r}=1,\\
\chi^{\ell}_{r}, & \text{if } m^{\ell}_{r}=0.
\end{cases}
\label{eq:ai_construction}
\end{equation}

This procedure preserves 50--70\% of the original response, retaining
student-specific linguistic and mathematical characteristics while
reducing obvious stylistic distinctions. Original responses are labeled
human-written, whereas partially completed and fully generated correct
responses form the AI-assisted class.

\paragraph{Detector Controls.}

Source human responses and their synthetic counterparts share a group
identifier and cannot cross data partitions. Detector inputs are
matched by item, score stratum, and length bin to reduce correctness-
and length-based shortcuts. The calibration temperature is fitted on
the validation set and then fixed. Generator transfer is evaluated by
withholding one generator family from detector development.

\subsection{AI-Aware Mastery-Target Weighting -- Methodology.}
The standard G-DINA model produces an unweighted mastery target
$c(s,k,t)$ for student $s$, principle $k$, and phase $t$. For AICG,
each response $r$ is assigned the reliability weight
\[
w_s^r=\max\!\left(\delta,1-p^{\mathrm{AI}}(s,r)\right),
\qquad \delta=0.30,
\]
where $p^{\mathrm{AI}}(s,r)$ is the calibrated AI-assistance
probability. Let $\ell_{s,r}(\boldsymbol{\zeta};\Theta)$ denote the
G-DINA log-likelihood contribution of response $r$ under
mastery profile $\boldsymbol{\zeta}$ and item parameters $\Theta$.
The reliability-weighted log-likelihood is
\[
J_w(s,t;\boldsymbol{\zeta})
=
\sum_{r\in R_s^t}
w_s^r\,
\ell_{s,r}(\boldsymbol{\zeta};\Theta).
\]
The corresponding weighted posterior is
\[
P_w(\boldsymbol{\zeta}\mid R_s^t)
=
\frac{
P_0(\boldsymbol{\zeta})
\exp\!\left[J_w(s,t;\boldsymbol{\zeta})\right]
}{
\sum_{\boldsymbol{\zeta}'}
P_0(\boldsymbol{\zeta}')
\exp\!\left[J_w(s,t;\boldsymbol{\zeta}')\right]
},
\]
and the AI-aware mastery target is its posterior marginal:
\[
\widetilde{c}(s,k,t)
=
\sum_{\boldsymbol{\zeta}}
\zeta_k
P_w(\boldsymbol{\zeta}\mid R_s^t).
\]
Finally, the reliability-adjusted learning-gain target is
\[
\widetilde{g}_{s,k}
=
\widetilde{c}(s,k,\mathrm{post})
-
\widetilde{c}(s,k,\mathrm{pre}).
\]
Full Cognitive uses the corresponding unweighted targets obtained by
setting all response weights to one. AI assistance therefore modifies
only the reliability of cognitive evidence; it does not alter the
instructor-assigned response-level grading labels.

\subsection{AICG Hyperparameter Settings -- Experiments} \label{app:hyperparameters} Table~\ref{tab:implementation_settings} reports implementation settings not specified in the main paper. The Gen-AI Detector and multi-task grading components both use \texttt{roberta-base}. Hyperparameters are fixed across all evaluation semesters and random seeds. \begin{table*}[!t]
\centering
\caption{Implementation settings for each component of AICG.}
\label{tab:implementation_settings} \small \setlength{\tabcolsep}{5pt} \begin{tabular}{lll} \toprule \textbf{Component} & \textbf{Setting} & \textbf{Final Value} \\ \midrule \multirow{8}{*}{Gen-AI Detector: contrastive pretraining} & Backbone & \texttt{roberta-base} \\ & Maximum sequence length & 512 tokens \\ & Projection dimension & 768 \\ & Balanced mini-batch size & 64 (32 per class) \\ & Optimizer & AdamW \\ & Learning rate & $2\times10^{-5}$ \\ & Training epochs & 10 \\ & Gradient-norm threshold & 1.0 \\ \midrule \multirow{7}{*}{Gen-AI Detector: linear classification} & Frozen components & Encoder and projection layer \\ & Classifier & 768-to-2 linear layer \\ & Maximum sequence length & 256 tokens \\ & Mini-batch size & 64 \\ & Optimizer & Adam \\ & Learning rate & $2\times10^{-4}$ \\ & Training epochs & 5 \\ \midrule \multirow{10}{*}{Multi-task Grading} & Backbone & \texttt{roberta-base} \\ & Maximum sequence length & 512 tokens \\ & Mini-batch size & One student record \\ & Optimizer & AdamW \\ & Learning rate & $2\times10^{-5}$ \\ & Weight decay & $0.01$ \\ & Training epochs & 8 \\ & Dropout probability & $0.10$ \\ & Learning-rate schedule & Linear decay \\ & Warmup proportion & $0.10$ \\ & Gradient-norm threshold & 1.0 \\ \midrule AICG evaluation & Random seeds & 42, 52, 62, 72, and 82 \\ \bottomrule \end{tabular} \end{table*} Mixed-precision training is used during supervised contrastive pretraining. The linear classifier is trained with standard cross-entropy while the pretrained encoder and projection layer remain frozen. Multi-task Grading is trained end to end with a linear learning-rate schedule following a warmup period covering the first 10\% of optimization steps.

\paragraph{G-DINA Configuration and Estimation.}
Pretest and posttest mastery profiles are estimated separately
using a fractional-response extension of the G-DINA model.
Each normalized instructor-assigned score
$y_{sr}\in[0,1]$ is used directly as fractional item-response
evidence rather than being converted into binary outcomes or
ordered response categories. For a mastery profile $\zeta$, the
contribution of response $r$ is
\[
y_{sr}\log \phi(s,r,\zeta;\Theta)
+
(1-y_{sr})\log\!\left[1-\phi(s,r,\zeta;\Theta)\right],
\]
where $\phi(s,r,\zeta;\Theta)$ is the G-DINA success probability.

Pretest and posttest parameters are estimated separately through
marginal maximum likelihood using the expectation--maximization
algorithm under an independent attribute distribution. We use
three random initializations, constrain success probabilities to
$[10^{-4},0.9999]$, and retain the solution with the highest
observed-data likelihood. Convergence requires the maximum
absolute changes in both item success probabilities and attribute
mixing proportions to fall below $10^{-4}$ between consecutive
iterations, with a maximum of 1,000 EM iterations.

For AICG, each fractional-response contribution is further scaled
by the reliability weight
$w_{sr}=\max(0.30,1-p^{\mathrm{AI}}_{sr})$, as defined in
Equation~(11). The resulting posterior mastery probabilities are
computed using Equation~(12). Full Cognitive uses the same
estimation procedure with $w_{sr}=1$ for every response.
All G-DINA parameters and mastery targets are estimated separately
within each expanding window using historical data only; the target
evaluation semester is excluded from both parameter estimation and
target construction.

\subsection{Baseline Configurations -- Experiments}
\label{app:baseline_configurations}

\paragraph{Common Evaluation Protocol.}
All baselines use the same expanding-window splits, preprocessing
pipeline, grade boundaries, and evaluation metrics as AICG. Transformer baselines receive the same textual input as AICG,
consisting of the question identifier, student response, and principle
metadata. ASQ-DBN follows its original answer, student, and question
feature construction, while prompt-based baselines receive the same
available information in textual form. No baseline
uses Q-matrix targets, cognitive labels, AI-assistance probabilities,
or reliability weights. Trainable baselines are evaluated with random
seeds $42$, $52$, $62$, $72$, and $82$, whereas prompt-based baselines
use five independent runs. Model selection uses only the historical
validation partition associated with each evaluation semester.

\paragraph{RoBERTa.}
RoBERTa is the score-only counterpart of AICG and uses
\texttt{roberta-base} with a sigmoid regression head
\cite{liu2019roberta}. To isolate the contribution of cognitive and
multi-objective supervision, it uses the same tokenization, maximum
sequence length of 512, student-record batching, and regression
objective as AICG. Specifically, the model is trained with
category-weighted Smooth L1 loss using $\beta=0.10$ and category
weights $(1.0,2.0,2.8,2.0,0.7)$. Optimization uses AdamW with a
learning rate of $2\times10^{-5}$, weight decay of $0.01$, dropout of
$0.10$, a linear warmup over the first 10\% of optimization steps, with gradient clipping at 1.0. Training lasts for eight epochs. Continuous
predictions are bounded to $[0,1]$, and discrete grades are obtained
using the common boundaries $0.125$, $0.375$, $0.625$, and $0.875$.

\paragraph{BERT.}
BERT replaces the RoBERTa encoder with
\texttt{bert-base-uncased} while retaining the same input, regression
head, loss, optimization settings, training epochs, and validation
procedure \cite{zhu2022bertasag}. This controlled adaptation isolates
the influence of the pretrained encoder without introducing additional
architectural or supervision differences. Its continuous and discrete
predictions are evaluated using the same procedures as those of
RoBERTa.

\paragraph{ASQ-DBN.}
ASQ-DBN follows the answer--student--question modeling principle of
\citeauthor{zhang2020going} \citeyear{zhang2020going}. The answer
representation is constructed using TF--IDF word unigrams and bigrams
with a maximum vocabulary of 10,000 terms, fitted only on the
historical training data. Student features comprise the mean, standard
deviation, and number of scores observed strictly before the evaluation
semester; unseen students receive the historical population statistics
and an additional missing-history indicator. Question features include
a one-hot question identifier and the historical mean, standard
deviation, and number of responses for that question. No score from the
evaluation semester is used to construct student or question features.

The concatenated answer, student, and question features are processed
by a deep belief network with hidden dimensions 512 and 128. Each layer
is initialized through 20 epochs of contrastive-divergence
pretraining, after which the complete network is fine-tuned using Adam
with a learning rate of $10^{-3}$, batch size 64, dropout of $0.20$,
and weight decay of $10^{-5}$. Fine-tuning runs for at most 50 epochs
with early stopping after five epochs without validation improvement.
A five-class softmax layer produces grade probabilities. The discrete
prediction is the highest-probability category, while its continuous
counterpart is calculated as the expected normalized category score,
$\sum_{d=0}^{4}(d/4)\pi_d$.

\paragraph{GPT-4o Scoring.}
GPT-4o Scoring uses the fixed model snapshot
\texttt{gpt-4o-2024-08-06} without parameter updates
\cite{openai2024gpt4o,mello2025automatic}. Each request contains the
question identifier, principle descriptions and weights, student
response, and definitions of the five ordered grade categories. The
model is instructed to return only one category in
${0,1,2,3,4}$. We use temperature $0$, top-$p=1$, and a maximum
output length of 16 tokens. The first valid standalone category is
parsed from the output. An invalid output triggers one deterministic
repair request using the same model and prompt context.

The predicted category is used directly for exact accuracy, Macro-F1,
and QWK. For continuous evaluation, category $d$ is mapped to
$d/4$. Five independent API runs are conducted using the same model
snapshot and fixed prompt. Neither in-context examples nor responses
from the evaluation semester are used.

\paragraph{GradeOpt.}
GradeOpt follows the Grader--Reflector--Refiner framework of
\citeauthor{chu2025gradeopt} \citeyear{chu2025gradeopt}. A separate
grading guideline is optimized for each question using only the
historical training and validation data. The initial guideline contains
the available principle metadata, principle weights, and five-category
grade definitions. GPT-4o is used for the Grader, Reflector, and
Refiner. The Grader uses temperature $0$, while the Reflector and
Refiner use temperature $0.5$.

Following the original configuration, guideline optimization uses an
outer batch size of 64, an inner batch size of 8, five outer iterations,
three inner iterations, and a UCB beam size of 4. Each optimization
batch contains equal proportions of high-error and randomly sampled
historical examples. Cohen's kappa on the historical validation set
serves as the guideline-selection criterion. Inner or outer
optimization terminates when no improvement is observed for two
consecutive iterations.

To preserve temporal isolation, the optimized guideline is frozen
before evaluation and GradeOpt's test-time adaptation is disabled.
Thus, no response or label from the evaluation semester is provided to
the Grader, Reflector, or Refiner during guideline optimization. The
final grader outputs one category in ${0,1,2,3,4}$, which is mapped
to $d/4$ for continuous evaluation. Five independent optimization and
evaluation runs are reported.

\end{document}